\documentclass[letterpaper, 10 pt, conference]{ieeeconf}  

\IEEEoverridecommandlockouts                              

\usepackage{amsmath} 
\usepackage{amssymb}  

\usepackage{graphicx}
\usepackage{comment}
\usepackage{caption}
\usepackage{verbatimbox}
\usepackage{stackengine}

\usepackage[hidelinks]{hyperref}
\usepackage[capitalise]{cleveref}
\crefname{equation}{}{}
\crefname{figure}{Fig.}{Figs.}

\usepackage{multicol}
\usepackage{subfigure}  
\makeatletter
\let\NAT@parse\undefined
\makeatother
\usepackage[square,numbers,sort&compress]{natbib} 

\title{\LARGE \bf
Optimizing Bipedal Locomotion for The 100m Dash \\ With Comparison to Human Running
}
\author{Devin Crowley, Jeremy Dao, Helei Duan, Kevin Green, Jonathan Hurst, and Alan Fern
\thanks{*This work is supported by the NSF Grant No. IIS-1849343 and DARPA Contract W911NF-16-1-0002.}
\thanks{All authors are with Collaborative Robotics and Intelligent Systems Institute, Oregon State University, Corvallis, Oregon, 97331, USA. }
\thanks{Email: \{\tt\footnotesize crowleyd, daoje, duanh, greenkev, jhurst, afern\}@oregonstate.edu.}
}

\begin{document}

\maketitle
\thispagestyle{empty}
\pagestyle{empty}

\begin{abstract}

In this paper, we explore the space of running gaits for the bipedal robot Cassie. Our first contribution is to present an approach for optimizing gait efficiency across a spectrum of speeds with the aim of enabling extremely high-speed running on hardware. This raises the question of how the resulting gaits compare to human running mechanics, which are known to be highly efficient in comparison to quadrupeds. 
Our second contribution is to conduct this comparison based on established human biomechanical studies. We find that despite morphological differences between Cassie and humans, key properties of the gaits are highly similar across a wide range of speeds. 
Finally, our third contribution is to integrate the optimized running gaits into a full controller that satisfies the rules of the real-world task of the 100m dash, including starting and stopping from a standing position. We demonstrate this controller on hardware to establish the Guinness World Record for \textit{Fastest 100m by a Bipedal Robot}.

\end{abstract}

\section{INTRODUCTION}
\label{sec:intro}

In recent years, reinforcement learning (RL) has proven highly effective for sim-to-real training of bipedal locomotion \cite{Li2021BerkeleyCassie, Castillo2022DigitRL,Taylor2021}. This has included learning to perform all bipedal gaits (e.g. walking, running, skipping, etc.) \cite{siekmann2020simtoreal}, blind stair climbing \cite{Siekmann2021Stairs}, carrying dynamic loads \cite{Dao2022UnsensedLoads}, and the first successful 5k run by a human-scale robot biped \cite{dao2021ICRA_wkshp_realDeployment}. There are existing works to enable running behaviors on bipedal robots, but they are limited by 2D approximations \cite{Cotton2012, Martin2015}, or the robot is specifically designed for running \cite{Ma2017}, whereas the Cassie robot can perform other types of gaits as well.

All of these bipedal locomotion behaviors were conducted at moderate speeds of up to $\sim2\frac{m}{s}$ on hardware. This raised the question of how fast a biped such as Cassie can run and how to develop a controller aimed at maintaining stability and efficiency across a range of speeds. The main goal of this paper is to address this question with an aim toward establishing the 100m dash world record for a bipedal robot. 

Prior work on learning for bipedal locomotion has included a number of fundamental gait parameters (e.g. frequency) and either assumed they were fixed \cite{siekmann2020simtoreal} or used a hand-tuned mapping from speed to gait parameters \cite{Dao2021MSProject}. To the best of our knowledge, no prior work has systematically explored how best to modulate the gait parameters across speeds for a real-world bipedal robot. 

Our first contribution is to explore the gait parameter space in a principled way in order to optimize gaits across a range of running speeds for the bipedal robot Cassie. The result is a speed-to-parameter curve that is qualitatively different from the current hand-tuned mapping. Our second contribution is to compare the optimized running mechanics of Cassie to human running mechanics, based on existing biomechanics literature. Interestingly, we find that there are compelling similarities despite the morphological differences between Cassie and humans. 

Finally, our third contribution is to integrate the optimized gaits into a full controller that satisfies the rules of a 100m dash. This includes the additional challenges of reliably starting from and stopping in a standing position. The resulting controller was used to establish the Guinness World Record for \textit{Fastest 100m Dash by a Bipedal Robot}.

\begin{figure}[t]
    \centering
    \includegraphics[width=\columnwidth]{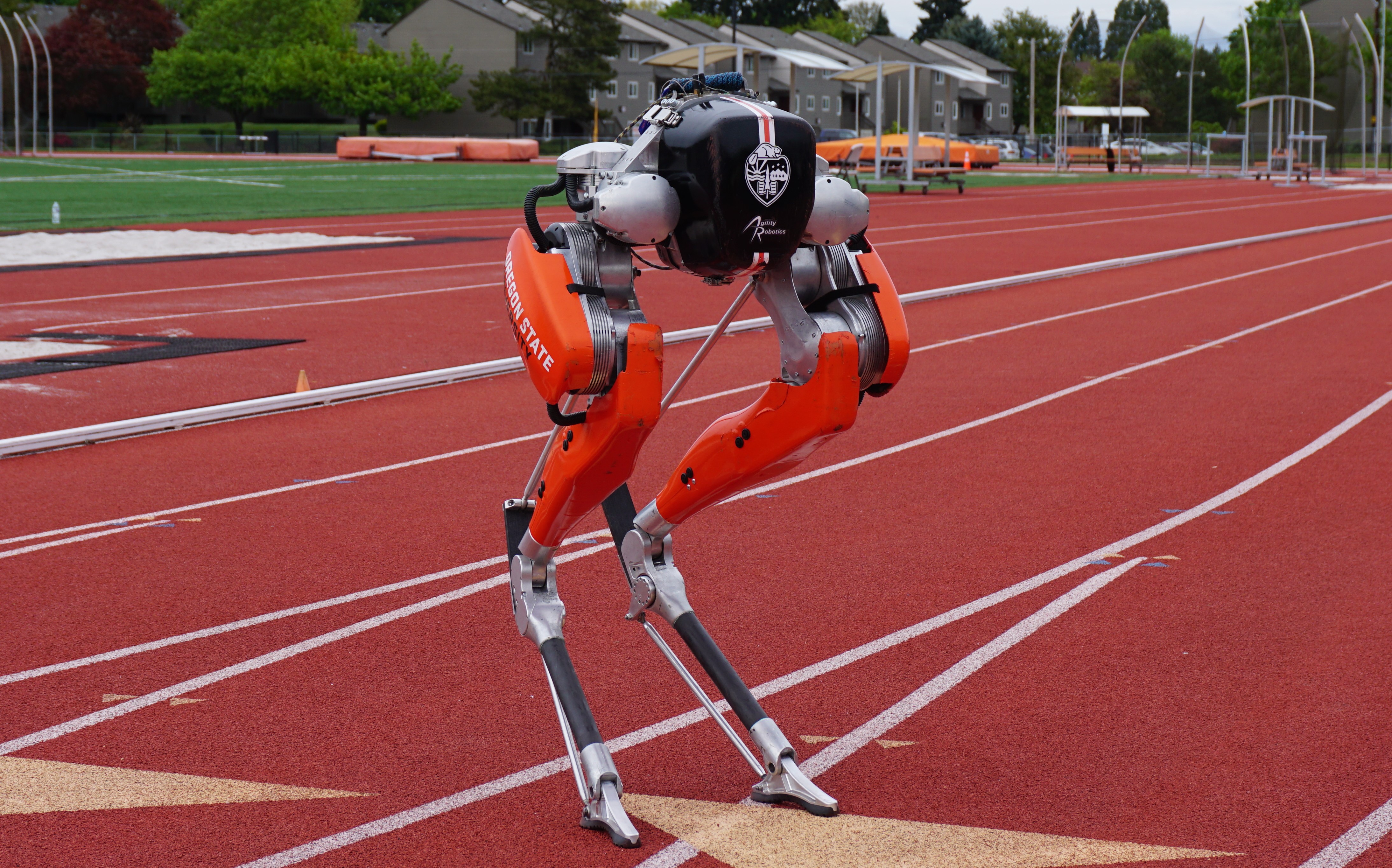}
    \caption{The bipedal robot Cassie standing on the track after setting the world record for \textit{Fastest 100m by a Bipedal Robot}.}
    \label{fig:cassie_on_track}
\end{figure}

\section{BACKGROUND}
\label{sec:background}

Our approach draws on prior work that uses sim-to-real reinforcement learning (RL) to train controllers for the bipedal Cassie robot, shown in Figure \ref{fig:cassie_on_track}. Below we summarize the controller architecture and RL approach that we adopt for this work. We also review how prior work selected running gait parameters, which we expand on in Section \ref{sec:optimizing gaits}.

{\bf Controller Architecture.} Following prior work \cite{siekmann2020simtoreal} we focus on blind locomotion, where the input to the running controller includes: 1) a 35-dimensional vector containing the joint positions and velocities for both actuated and unactuated joints, as well as the pelvis orientation and rotational velocity, 2) a clock signal (the phase) that encodes a position within the gait cycle and guides the swing versus stance phases of the gait, 3) the gait parameters $freq$ and $ratio$ (defined later in this section), and 4) the set of external commands to the robot, which for our straightline running goal consists of only the target forward velocity.
The controller outputs are the PD controller targets for the 10 actuated joints (5 per leg) centered around a nominal standing pose. The learned running controller, or policy, outputs PD targets at 40Hz, while the low-level PD controllers operate at 2kHz with fixed gains. This general approach of training a slower high-level controller to direct fast PD controllers is widely used for RL-based robot locomotion, e.g. \cite{peng2017learning,xie2018feedback, tan2018AgileLocomotion, hwangbo2019learning, TsounisDeepGait}. 
Our running controller is represented as a recurrent LSTM neural network. Its ability to augment the state with memory helps with the sim-to-real transfer.

{\bf Sim-to-Real Training.} The controller is trained via sim-to-real RL using the standard actor-critic PPO algorithm \cite{schulman2017proximal} with the ``clipped objective" variation. Training is conducted within the MuJoCo physics engine \cite{todorov2012mujoco} using a model of the Cassie robot and dynamics randomization of physical parameters such as friction and the center of mass. The resulting controller is directly deployed on the real robot. Dynamics randomization and memory-based recurrent network controllers have been shown to be a powerful combination for transferring learned controllers to the real world \cite{siekmann2020learning, lee2020anymalchallengeterrain, peng2018dynrand}. 

The above training framework can optimize for different gaits via appropriate specification of the reward function. The reward function produces a reward at each step of the controller (each policy step, i.e. at 40Hz), and we use the dense reward formulation from recent work \cite{Dao2021MSProject}. Intuitively, each gait is a parameterization of certain reward components that are aligned with the clock. For example, when a foot that is supposed to be in stance has zero velocity, we provide a positive reward. Further details can be found in \cite{Dao2021MSProject}.

For the purposes of training running policies in this work, the key gait parameters that govern the reward are: 1) the swing ratio ($ratio$), which is the proportion of time that the reward encourages each foot to be in the air (in swing) as opposed to planted on the ground (in stance), and 2) the stride frequency ($freq$), which modulates the number of footsteps per second. The reward function is defined such that controllers whose gaits better match the specified $ratio$ and $freq$ parameters achieve higher total reward. 

\begin{figure}[tb]
    \vspace{1mm}
    \centering
    \includegraphics[width=\columnwidth]{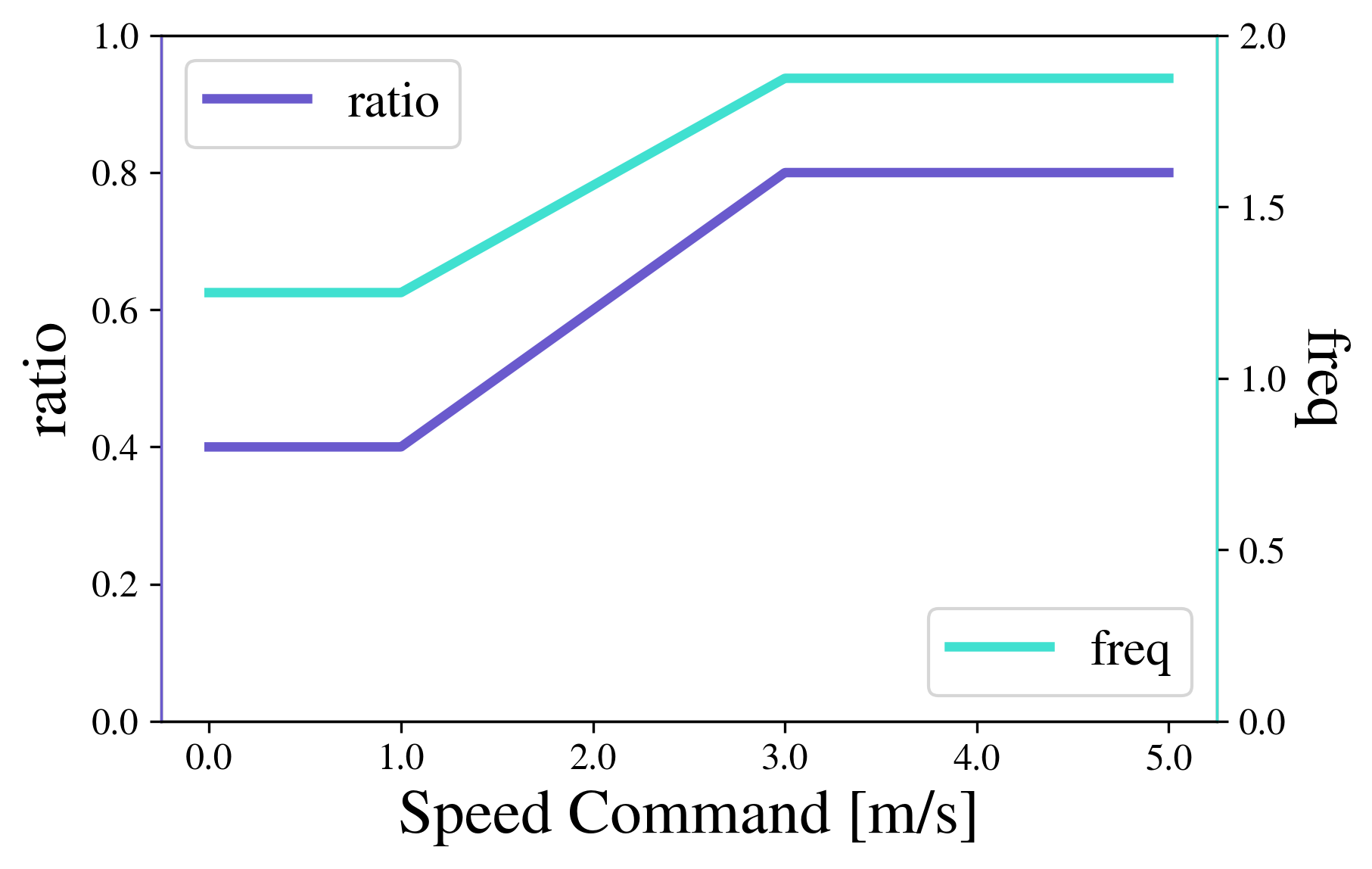}
    \caption{Hand-tuned mapping from the speed command to the swing ratio and stride frequency gait parameters.}
    \label{fig:heuristic}
\end{figure}

Selecting appropriate gait parameter values for training has remained a challenge. Previous work \cite{siekmann2020simtoreal} trained Cassie using fixed user-specified values that did not change with the current speed command during training. However, it is clear that these parameters should depend on the target velocity.

In recent related work for quadruped robots \cite{yang2022fast}, gait parameters were learned for a model-based controller, resulting in natural gait transitions. However, we are unaware of analogous work for bipedal robots, particularly attempting high speed running. \cite{Dao2021MSProject} developed a hand-tuned piecewise-linear mapping from the speed command to $ratio$ and $freq$ for Cassie, as illustrated in Figure \ref{fig:heuristic}. This mapping proved sufficient to produce robust walking and running behaviors. However, there is currently no theoretical or empirical rationale for the mapping.  One of the key contributions of our work is to provide a principled approach for such a mapping as developed in the next section.

\section{OPTIMIZING GAIT PARAMETERS}
\label{sec:optimizing gaits}

This section presents our approach to optimizing gaits across a range of speeds. Subsection A details how we train a policy on a spectrum of gait parameter values and speeds. Subsection B discusses how we select the best gait parameter values at a given speed.

\subsection{Training Across a Spectrum of Gait Parameter Values}

To learn a mapping from the speed command to $ratio$ \& $freq$ we first train a policy on all combinations of $ratio$ \& $freq$ within constant offsets from the hand-tuned mapping in Figure \ref{fig:heuristic}, across speeds of $0$ to $5\frac{m}{s}$. That is, for a given speed command, the training procedure chooses values for these gait parameters from uniform distributions respectively within offsets of $0.2$ and $0.625$ of the hand-tuned mapping for that speed command. 

Given the limited range of values in the hand-tuned mapping ($0.4$ to $0.8$ for $ratio$ and $1.25$ to $1.875$ for $freq$) these are fairly large maximum offsets, especially for $ratio$ which is intrinsically bound to $(0, 1)$. We find no lapse in performance from the size of these offsets.

\subsection{Learning a Mapping from Speed to Gait Parameters}

This policy is able to perform all gaits it was trained on, i.e. within the maximum offsets of the hand-tuned mapping. We ran a search in simulation to determine the best $ratio$ and $freq$ values at each speed in a discretization of the space. For each combination of gait parameters at a given speed, the robot takes 50 to 250 policy steps (based on the speed command) in simulation to accelerate to that speed and enter a stable locomotion pattern. These rollouts were deterministic and used the same initial state.

We collect trajectories of several cost terms 
for 100 policy steps after Cassie settles into the commanded speed. 
These costs are compiled into an overall score for each gait parameter combination at each speed. As a note, any simulations in which Cassie falls are scored as 0. The scoring metric we use is based on the following four costs, computed at each policy step:
\begin{itemize}
    \item Speed error: how well the speed command was matched
    \item Cost of transport: the energy used / distance traveled
    \item Torque cost: the average torque used
    \item Motor velocity: the average motor speed
\end{itemize}
Each component is weighted to have a similar contribution to the overall score across all speeds.

However, 3 of the 4 components measure efficiency whereas only the speed error cost measures performance. By giving each component an equal contribution, the overall score prioritizes efficiency over speed fidelity. Despite this, Cassie's speed error averaged $<0.2\frac{m}{s}$ for most gait parameter combinations, and even less for the top 5. In other words, Cassie is able to closely match its commanded speed for most gait parameter values but only the top scoring combinations are the most efficient.

\begin{figure}[tb]
    \vspace{1mm}
    \centering
    \includegraphics[width=\columnwidth]{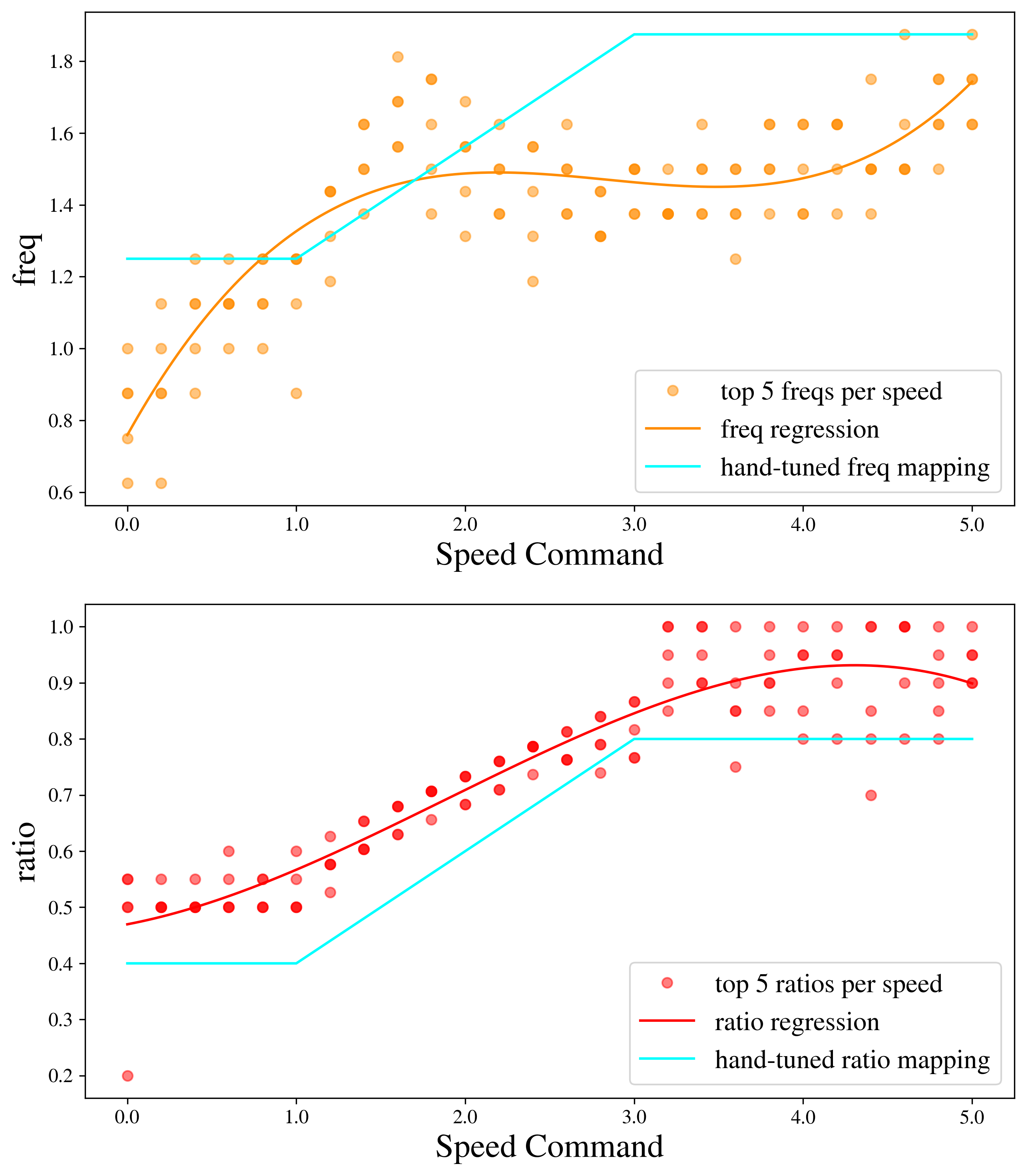}
    \caption{The top 5 most efficient $freq$ (above) and $ratio$ (below) gait parameter commands at each commanded speed. The blue line indicates the previous hand-tuned mapping, and the curve is a 3rd degree polynomial regression.}
    \label{fig:top 5 gait parameters}
\end{figure}

The rest of this exploration focuses on the top 5 gait parameter combinations at each speed command, as chosen by this scoring metric. Figure \ref{fig:top 5 gait parameters} shows the $freq$ and $ratio$ values of these top 5 combinations for each speed, plotted against the hand-tuned mapping in blue. 

We hypothesized that higher speeds are best enabled by high stride frequencies. However, 
these results prefer much lower stride frequencies than anticipated. This holds even when the scoring metric exclusively prioritizes speed fidelity over efficiency. Figure \ref{fig:top 5 gait parameters} shows that the best scoring $freq$ values decrease and hold steady between speeds of $2$ to $4\frac{m}{s}$. They increase again after $ratio$ tops out at $4\frac{m}{s}$. 
The $ratio$ prefers to increases to near maximum and plateau.

A lower $freq$ increases stride-time (the sum of swing-time and stance-time), whereas a higher $ratio$ shifts the distribution of stride-time towards swing-time, at the expense of stance-time. 
In other words, as Cassie approaches its top speed, these high scoring gait parameters prefer long, infrequent steps. The only way to accomplish this is with a long stride length, enabled by a significant aerial phase in which both feet are in swing (and neither in stance). The increased $ratio$ adds to the aerial time while the decreased $freq$ mitigates the corresponding decrease in stance-time. To decrease the stance-time by too much would require excessive forces during stance, 
since the feet must deliver enough impulse to maintain speed. 
The hand-tuned mapping used too high a $freq$ to allow long enough stance times, resulting in greater ground forces for the same speed.

\section{COMPARISON TO HUMAN RUNNING MECHANICS}
\label{sec:comparison}

Cassie's mechanics are similar to those of bipedal animals. This section explores how similar its emergent behavior is to human running mechanics when we optimize its gait.
We measure various running mechanics for each of the top 5 scoring gait parameter combinations at each speed to compare against previously published running mechanics for humans in the same body-agnostic units.

\begin{figure*}[tb]
    \centering
    \subfigure{\includegraphics[height=0.4\textheight]{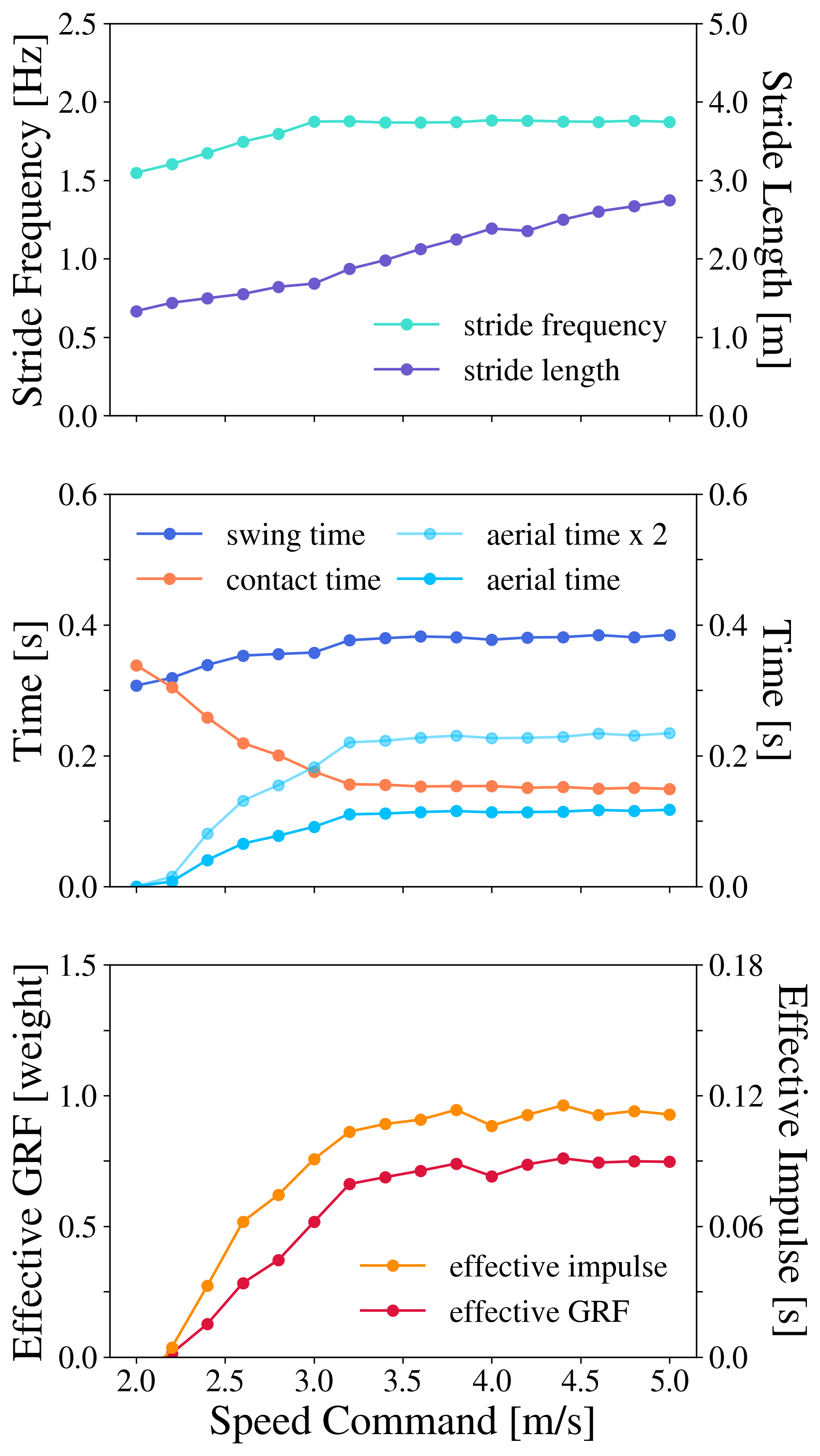}}
    \hfill
    \subfigure{\includegraphics[height=0.4\textheight]{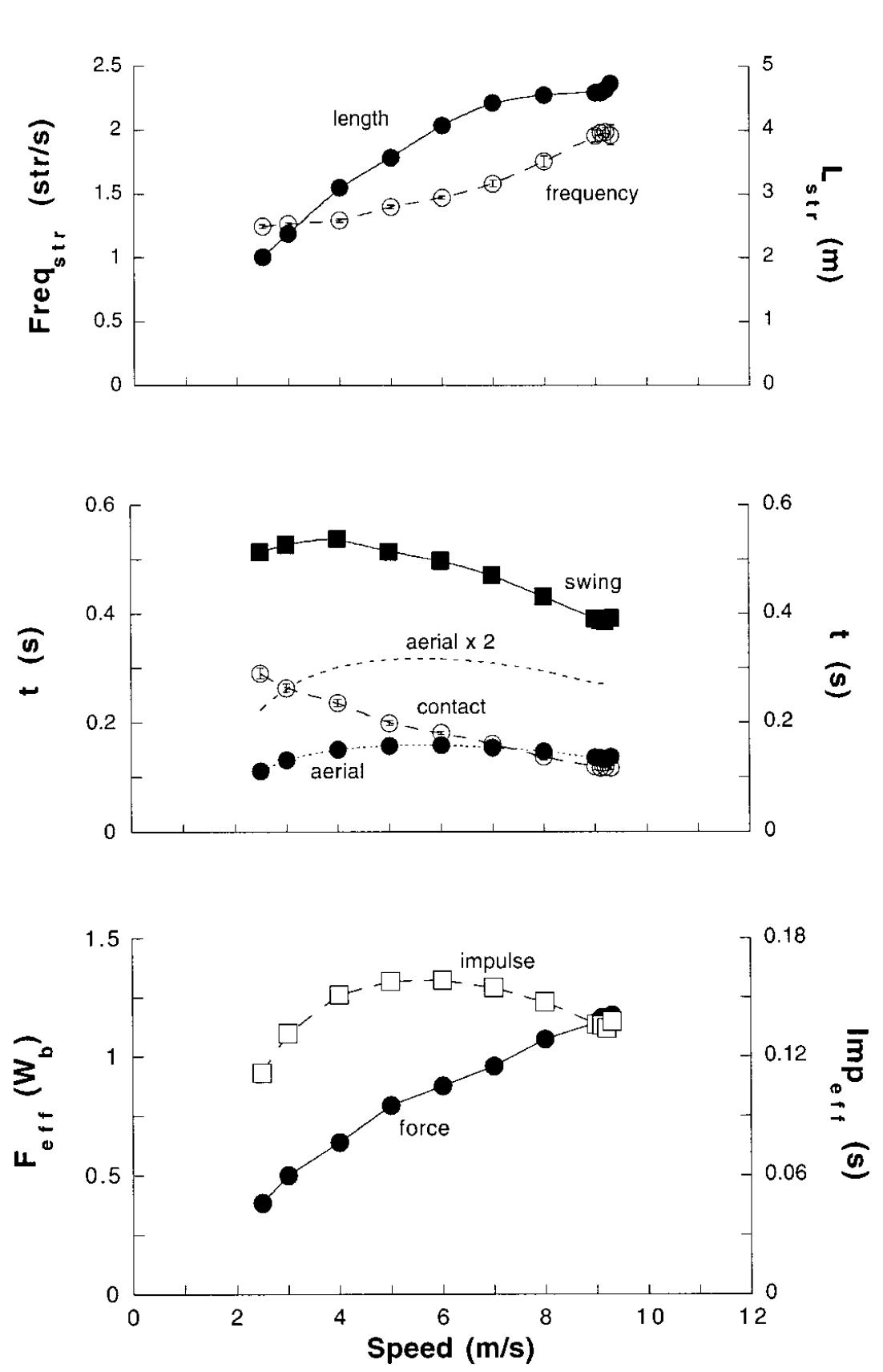}}
    \hfill
    \subfigure{\includegraphics[height=0.4\textheight]{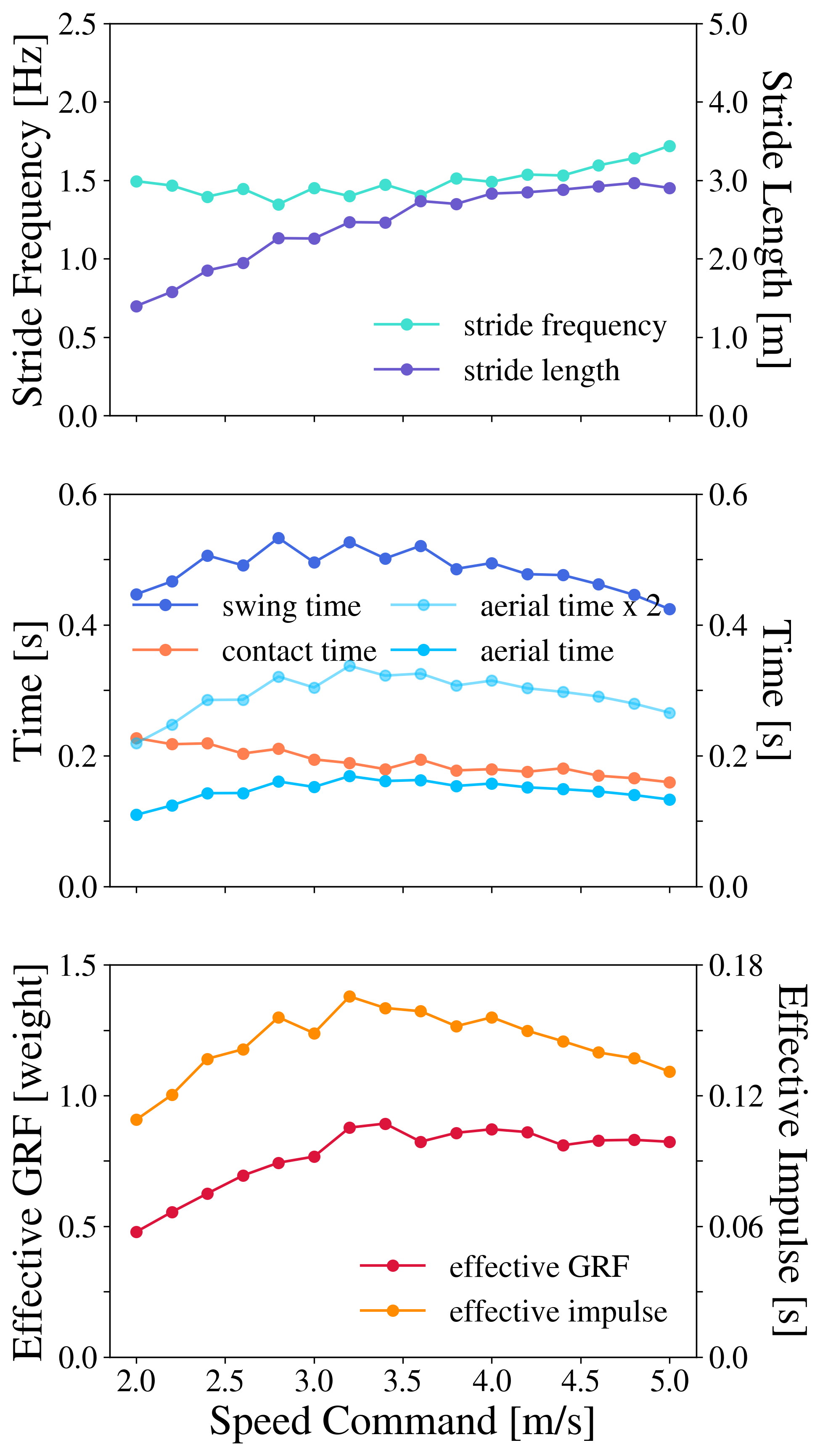}}
    \vspace{-3mm}
    \begin{multicols}{3}
        \textbf{
            \hspace{-5mm}Hand-Tuned\\
            Humans\\
            Optimized\\
        }
    \end{multicols}
    \vspace{-3mm}
    \caption{Running mechanics as functions of speed (left) for the hand-tuned gait parameter mapping with Cassie, (center) as measured on humans, (right) for Cassie averaged across the top 5 most efficient gait parameter values at each speed. The center plot for humans is taken directly from \cite{weyand2000faster}.}
    \label{fig:running mechanics}
\end{figure*}

The center column of Figure \ref{fig:running mechanics} (humans) is a series of plots from \cite{weyand2000faster} showing several running mechanics as functions of speed measured on humans. From the top plot we see humans significantly increase their stride length with increased speed, tapering at high speeds. They show an inverse behavior for stride frequency.
This means that until the stride length begins to taper, increases in speed are primarily brought about by increased stride length rather than increased stride frequency, but this reverses for top speeds.

In the middle plot of the human column of Figure \ref{fig:running mechanics} we see that the swing-time decreases at moderate to high speeds, and the aerial time constitutes $\sim\frac{2}{3}$ of the total swing-time. To see this we must compare ``aerial x 2" to the swing-time since each foot experiences 2 aerial phases per swing phase. When running, think of the right foot touching down as splitting the left foot's swing phase into two aerial phases.

In the lower plot of the human column of Figure \ref{fig:running mechanics} we see a near-linear increase in the effective ground reaction force ($GRF_{eff}$) as a function of speed, while the effective impulse -- the product of $GRF_{eff}$ with stance-time, or contact-time -- is maximized at moderate speeds before decreasing again. This observation about $GRF_{eff}$ is the core insight of \cite{weyand2000faster}, the paper that the human column comes from. The insight is in its title: \textit{Faster Top Running Speeds Are Achieved With Greater Ground Forces Not More Rapid Leg Movements}.

Effective ground reaction force is defined as:
\begin{align}
    GRF_{eff} &= \frac{GRF - F_{weight}}{F_{weight}}
    \label{eqn:grf_effective}
\end{align}
This definition of $GRF_{eff}$ translates to body-agnostic vertical acceleration and directly affects aerial time \cite{weyand2000faster}.

The right column in Figure \ref{fig:running mechanics} (optimized) is a recreation of the human column using averages from the top-5-scoring gait parameter combinations at each speed. We present Cassie's running mechanics starting at $2\frac{m}{s}$ rather than $0\frac{m}{s}$ in accordance with the human column and because this is the speed at which Cassie develops an aerial phase and can be considered running.

In the optimized column of Figure \ref{fig:running mechanics}, the top plot shows very similar behavior to its counterpart in the human column. The stride length increases with a downward concavity and tapers at top speeds, and the stride frequency is mostly flat but increases at top speeds. The stride frequency has a small discrepancy compared to the human measurements at lower speeds in that it briefly decreases, but the concavities are in full agreement. Not only are the qualitative behaviors in accord but the actual values show similar ranges as well. The human column covers speeds on the abscisssa from $\sim2.5$ to $9.5\frac{m}{s}$ whereas the optimized column covers speeds from $2$ to $5\frac{m}{s}$, since Cassie is still slower than adept human runners.

The middle plot of the optimized column of Figure \ref{fig:running mechanics} is again in strong agreement with its counterpart in the human column; however, the lower plots show somewhat different behavior. The effective impulse is maximized at moderate speeds in both plots, but $GRF_{effective}$ in the optimized column reaches a limit and tapers after $\sim3\frac{m}{s}$ whereas it increases $\sim$linearly at all speeds in the human column.

To explain this difference we found that the hip-roll motors reached maximum torque above $\sim3\frac{m}{s}$. We hypothesize that this limits the maximum $GRF_{effective}$ and that greater forces would cause instability in the pelvis, which we disincentivize in our reward function.
Despite this difference, the strong similarities that emerge are compelling. 
In particular, the running mechanics in the optimized column of Figure \ref{fig:running mechanics} are qualitatively much closer to those in the human column than in the hand-tuned column. 
This gives us confidence that using a learned mapping from speed to gait parameters will lead to more natural and efficient locomotive behaviors.

\section{100m DASH CONTROLLER}
\label{sec:dash controller}

In this section we lay out the requirements for the 100m dash and discuss the optimizations and adjustments made to the controller for this task. These rules are consistent with those from the Guinness World Record \textit{Fastest 100m Dash By a Bipedal Robot}, which we have established and now hold with Cassie.

We break the 100m dash into a timeline with 5 stages, which encapsulate the key requirements of the official record:
\begin{enumerate}
    \item Robot stands, both feet at rest behind the start line.
    \item Time starts. Robot accelerates to its top speed.
    \item Robot runs steadily at top speed towards the finish line.
    \item Time stops. Robot slows until it is stepping in place, having crossed the finish line.
    \item Robot plants its feet, returning to a static standing pose.
\end{enumerate}
All 5 stages must be completed without falling.

There is significant space to optimize for this task, particularly in stages 2 \& 3. Stages 1, 4, \& 5 are essentially requirements for the record, but so long as they are successfully accomplished, the degree of that success has no bearing on the final time and therefore the quality of the record.

\subsection{Specialization}

We make several adjustments to Cassie's controller to specialize it to the 100m dash. We train it only on non-negative speeds, and up to $5\frac{m}{s}$. We also reduce its command set to the bare minimum: forward speed. In doing so we remove the commands for strafing and turning from the controller's observed state and eliminate them from training. This is to say that Cassie is only asked to run straight forward in training.

As a neural network, the running policy has limited representational power to encode behavior. By stripping away those behaviors that are unnecessary for the 100m dash we specialize the policy. As a result, this policy specialized for running is unable to strike a static standing pose. The closest that it can do is step in place when the speed command is $0\frac{m}{s}$. For this reason we use two policies: our running policy for stages 2, 3, \& 4, and a pretrained standing policy whose only capability is standing still for stages 1 \& 5. 

One optimization we do not pursue is using a separate specialized policy for stage 2, the takeoff and speed ramp up. Cassie accelerates to its top speed in only a few seconds so the vast majority of measured time is spent in stage 3. We therefore deem it acceptable to focus on a single running policy rather than separate ``speedup" and ``steady-state" running policies.

Using two policies introduces the challenge of transitioning between them. We take the simplest approach. Rather than learning the transitions explicitly, we swap out the controlling policy at the right moment. The rest of this section discusses finding the right moment.

\subsection{Taking the First Step: Standing to Stepping}

\begin{figure}[tb]
    \vspace{1mm}
    \centering
    \includegraphics[width=\columnwidth]{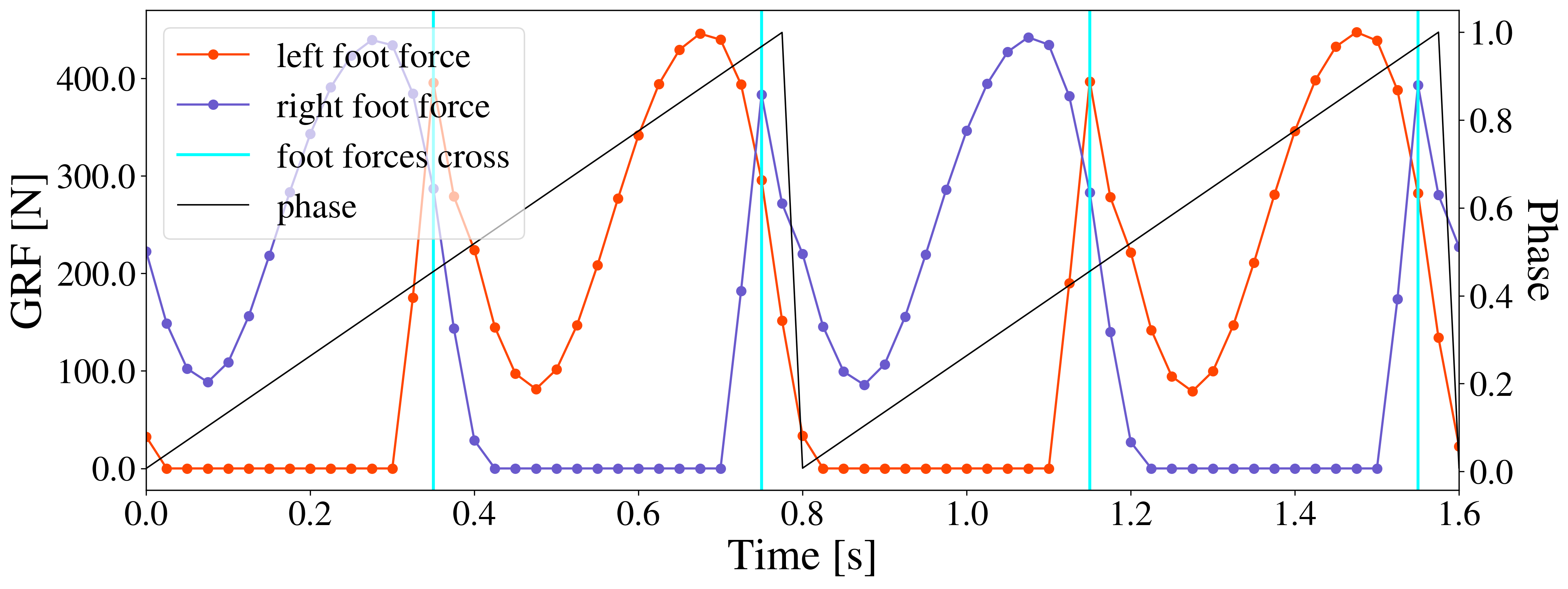}
    \caption{The ground reaction force (GRF) at each policy step together with the phase. This shows 2 gait cycles and emphasizes the phases at which these forces cross with vertical lines. These are the best phases for transitioning from standing to stepping.}
    \label{fig:foot forces}
\end{figure}

The first policy transition occurs between stages 1 \& 2, as Cassie takes its first step. The running policy takes a cyclic clock signal as part of its input; we call this the phase. It encodes where Cassie should be within its gait cycle, i.e. which feet should be in swing versus in stance. For this transition we need to choose what phase to start at.

Figure \ref{fig:foot forces} shows the ground reaction force (GRF) for each foot as a function of time along with the phase. 
This is for a running policy stepping in place. Note that whenever the GRF for a foot is nonzero, that foot is in stance. It is best to make this transition at the phase where Cassie's stepping behavior is most similar to its standing behavior: when both feet are at rest, pressing down. These moments are when the GRFs for both feet are as similar as possible. We identify these moments as the phases when the GRFs have just crossed, marked by vertical lines in Figure \ref{fig:foot forces}.

There is some expected variability in these GRF crossing phases but the distribution is highly bimodal and tightly clustered. This bimodality is because there are two moments in each gait cycle when the GRFs cross, one as each foot touches down. 
Our controller sets the phase to a static value as it transitions from the standing policy to the running policy. 
For this value we ran a running policy in simulation for 1200 policy steps, or $\sim$30 footsteps, and arbitrarily chose the median GRF crossing phase.

\subsection{Taking the Last Step: Stepping to Standing}

\begin{figure}[tb]
    \vspace{1mm}
    \centering
    \includegraphics[width=\columnwidth]{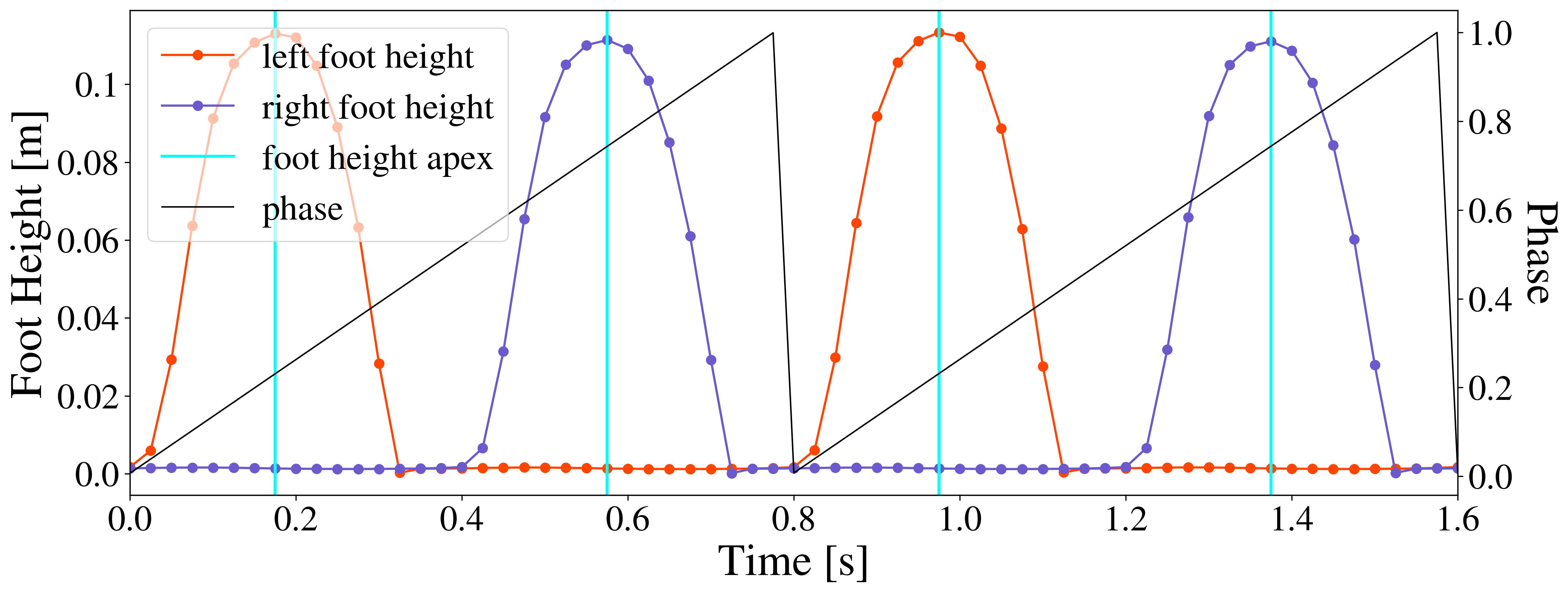}
    \caption{Foot heights across 2 gait cycles. To allow robust switching from stepping to standing the controller waits for an apex phase, marked by the vertical lines.
    }
    \label{fig:foot heights}
\end{figure}

The transition in stage 5 is more difficult because the standing policy is less stable than the running policy. The running policy is able to freely take steps to correct imbalance -- it does so constantly. The standing policy 
is specifically encouraged to maintain balance while keeping both feet planted and so it has much less practice taking steps and tends to wait until it's too late.

The standing policy is trained to plant its feet and balance starting from a suite of dynamic poses taken from trajectories of a running policy stepping in place or walking at low speed. While the standing policy is effectively unable to lift its feet and step, it is able to plant a foot that is already in the air and position it. If the transition occurs when both feet are already planted then the standing policy can only hold them in place and try to balance, but with the feet planted too close together it has no way to adequately absorb momentum causing Cassie to tip over and fall.

Unlike in the transition from standing to stepping we cannot immediately swap to the new policy and set the phase -- the standing policy has no sense of phase to begin with. Instead when the controller receives the signal to stand it waits until the phase is in a good position in the gait cycle and then swaps to the standing policy. The best moment in the gait cycle is when either foot is at its apex, giving the standing policy the greatest opportunity to position that foot in the direction of Cassie's momentum and create a wide standing base.

Figure \ref{fig:foot heights} shows the foot heights as a function of time along with the phase for a running policy stepping in place. 
The moments when a foot is at its highest point are marked by vertical lines. There are two phases per gait cycle when one foot is at its apex, depending on which foot. Either of these phases is a good moment to transition.

As before, we choose the medians of these apex phases for each foot as the transition points. 
When Cassie receives the signal to transition from stepping to standing, the controller waits for either of these two phases and swaps policies at that moment.

We make one other adjustment to achieve consistent transitions from stepping to standing. In addition to transitioning when either foot is at its apex, it is necessary to reduce the speed command below zero to a small negative value. The running policy tends to walk forward slightly when commanded $0\frac{m}{s}$ since most of its training is for running and accelerating forward. Fortunately it is able to extrapolate to this negative speed command and step in place.

Prior to these adjustments the controller made the transition from stepping to standing immediately when given the signal to swap. This setup was severely unreliable and almost always failed, both on hardware and in simulation (less than $10\%$ success rate). If we sent the signal to swap back to the running policy after seeing the standing policy begin to tip over it could usually recover with the running policy's superior stability, but this left us back where we started. Success depended heavily on the timing, which was done manually. After making the controller wait for an apex phase the success rate soared to about 2 in 3 attempts, and with this final adjustment to the speed command it improved beyond expectations to $100\%$ of $>$20 trials.

\section{100M DASH RESULTS}
\label{sec:results}

\begin{table}[tb]
    \caption{100m Dash Trials}
    \begin{center}
    \addstackgap[0mm]{
        \begin{tabular}{|c|c|c||c|c|c|}
            \hline
            \multicolumn{3}{|c||}{Hardware} & \multicolumn{3}{|c|}{Simulation}\\
            \hline
            Trial & Time [s] & Speed [m/s] & Trial & Time [s] & Speed [m/s]\\
            \hline
            1 & 26.11 & 3.83 & 1 & 24.53 & 4.08\\
            2 & 24.37 & 4.10 & 2 & 24.38 & 4.10\\
            3 & 27.38 & 3.65 & 3 & 24.43 & 4.09\\
            \hline
        \end{tabular}
    }
    \end{center}
\vspace{-5mm}
\label{tab:trial_times}
\end{table}

Our official 100m dash attempts for the world record were conducted on May 11, 2022, at the Whyte Track and Field Center in Corvallis, Oregon. Cassie fell in the first 5 seconds of 2 early trials due to operator error, but following these were 3 successful trials without falling. We invited two independent witnesses with expertise in robotics and two independent timekeepers to measure the final time on stopwatches. One witness signalled the start of each dash. Our official world record dash is presented in our supplementary video.

\begin{figure}[tb]
    \vspace{1mm}
    \centering
    \includegraphics[width=\columnwidth]{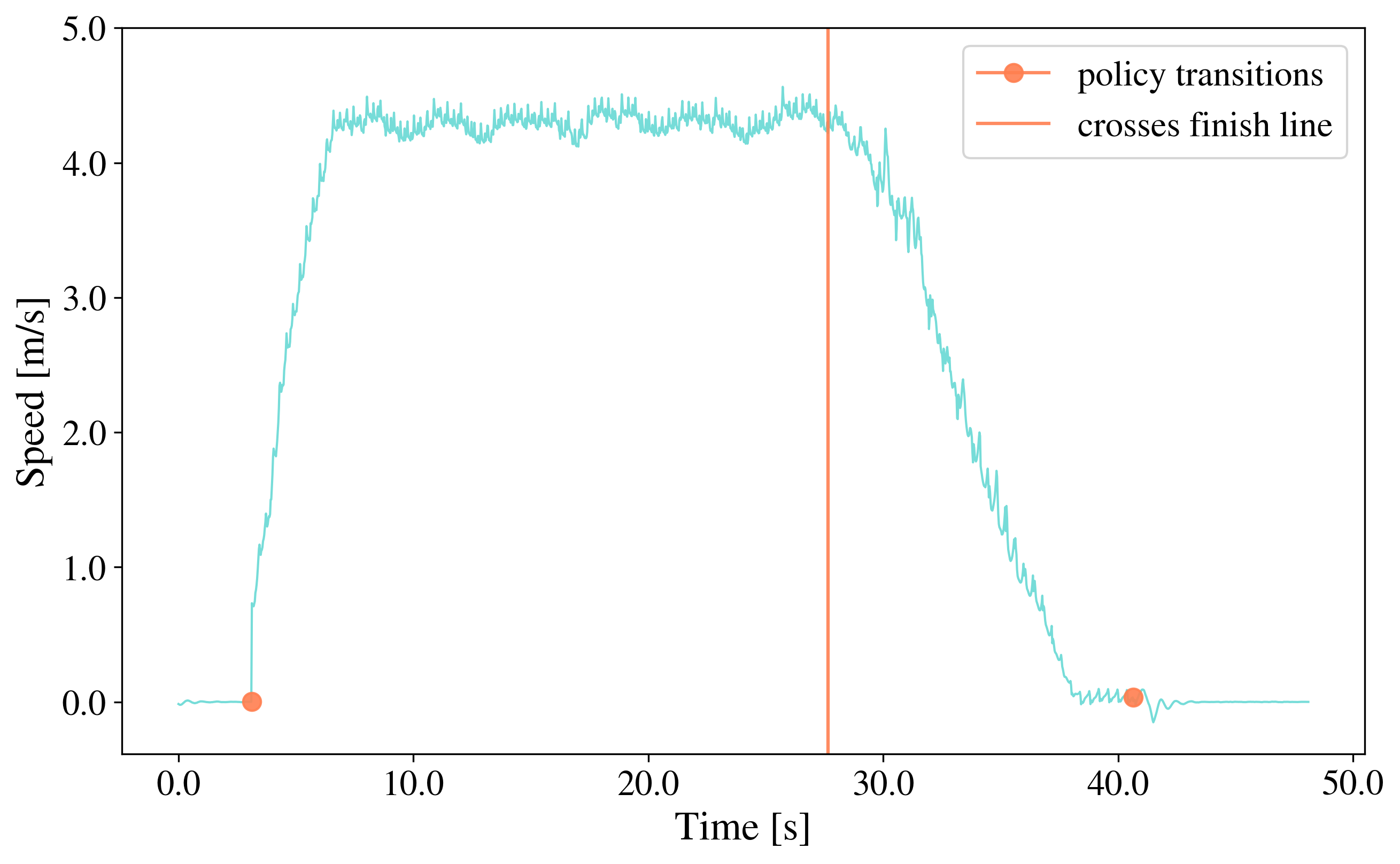}
    \caption{Speed as a function of time for a simulated 100m dash. The dots mark the policy transitions between the standing and running policy, and the vertical line marks the moment when Cassie crosses the finish line.}
    \label{fig:simulation speeds}
\end{figure}

The times and average speeds of our official dashes are presented in the left half of Table I. The right half presents results from complementary trials in simulation using the same controller and operating procedure, including the same speed command of $4\frac{m}{s}$. 
Figure \ref{fig:simulation speeds} shows the speed profile of one such simulated trial. 
The simulated dash times were tightly clustered very close to the best real time. The similarity is closer than anticipated but the spread on hardware is to be expected, since the variation primarily stemmed from user error in adjusting the orientation.

However, in simulation we can increase the speed command up to $5\frac{m}{s}$ and have Cassie achieve that speed without losing stability and falling. This is not possible on hardware with the present controller. 
The sim-to-real gap, arising from 
inaccuracies in the simulated model, causes this difference in performance.
Taking advantage of this gap and operating the controller at $5\frac{m}{s}$ yields 100m dash times of $\sim$22 seconds, at an average speed of over $4.5\frac{m}{s}$.
Previous running controllers for Cassie have shown similar performance in simulation, but none have been bridged the gap onto hardware so effectively.

\section{CONCLUSION}
\label{sec:conclusion}

In this paper we have presented a systematic approach to gait optimization across a spectrum of speeds for bipedal running using the robot Cassie. We compared the optimized gaits to human running mechanics and found compelling parallels. Finally, we integrated these optimized gaits into a full controller specialized for a 100m dash and established the Guinness World Record for \textit{Fastest 100m Dash by a Bipedal Robot}. Interestingly, the 100m dash controller was trained on a wide spectrum of gait parameters and then operated using the optimized parameters. This leaves open the possibility of further improvements by exclusively training on the optimized gait parameters. 
The current speeds we have achieved on Cassie are over $4\frac{m}{s}$, roughly $\frac{1}{3}$ those of top human runners. Closing this gap via a combination of hardware and software innovations remains a compelling challenge problem for the bipedal robotics community. 

\newpage 
\bibliographystyle{IEEEtran.bst}


\end{document}